%% file: main.tex
\documentclass[10pt,twocolumn,letterpaper]{article}

\usepackage[pagenumbers]{cvpr} 
\usepackage[accsupp]{axessibility}  

\input{preamble}

\definecolor{cvprblue}{rgb}{0.21,0.49,0.74}
\usepackage[pagebackref,breaklinks,colorlinks,citecolor=cvprblue]{hyperref}

\def\paperID{11080} 
\def\confName{CVPR}
\def\confYear{2025}

\title{Post-pre-training for Modality Alignment in Vision-Language Foundation Models}

\author{%
  Shin'ya Yamaguchi\thanks{Corresponding author. \texttt{shinya.yamaguchi@ntt.com}}\\
  NTT, Kyoto University\\
  \and
  Dewei Feng \\
  MIT\\
  \and  
  Sekitoshi Kanai \\
  NTT \\
  \and
  Kazuki Adachi \\
  NTT, YNU\\
  \and
  Daiki Chijiwa \\
  NTT\\
}

\begin{document}

\maketitle

\begin{abstract}
Contrastive language image pre-training (CLIP) is an essential component of building modern vision-language foundation models.
While CLIP demonstrates remarkable zero-shot performance on downstream tasks, the multi-modal feature spaces still suffer from a modality gap, which is a gap between image and text feature clusters and limits downstream task performance.
Although existing works attempt to address the modality gap by modifying pre-training or fine-tuning, they struggle with heavy training costs with large datasets or degradations of zero-shot performance.
This paper presents CLIP-Refine, a post-pre-training method for CLIP models at a phase between pre-training and fine-tuning.
CLIP-Refine aims to align the feature space with 1 epoch training on small image-text datasets without zero-shot performance degradations.
To this end, we introduce two techniques: random feature alignment (RaFA) and hybrid contrastive-distillation (HyCD).
RaFA aligns the image and text features to follow a shared prior distribution by minimizing the distance to random reference vectors sampled from the prior.
HyCD updates the model with hybrid soft labels generated by combining ground-truth image-text pair labels and outputs from the pre-trained CLIP model.
This contributes to achieving both maintaining the past knowledge and learning new knowledge to align features.
Our extensive experiments with multiple classification and retrieval tasks show that CLIP-Refine succeeds in mitigating the modality gap and improving the zero-shot performance\footnote{Code: \url{https://github.com/yshinya6/clip-refine}}.

\end{abstract}

\section{Introduction}\label{sec:intro}
Contrastive language image pre-training (CLIP, \cite{Radford_ICML21_CLIP,Jia_ICML21_ALIGN_contrastive}) is a standard method to build modern vision-language foundation models.
CLIP enables models to learn multimodal representations to map images and texts into a shared feature space with contrastive loss on large-scale image-text pair datasets.
Since the pre-trained CLIP models provide a cross-modal understanding of input image/text data in various domains, they are widely used as a foundation of many applications, including zero-shot classification~\cite{Radford_ICML21_CLIP,Ge_CVPR23_improving_zeroshot_clip_by_hierarchy,Wang_ICCV23_improving_zeroshot_clip_by_synthetic_caption}, cross-modal retrieval~\cite{Iscen_ICLR24_retrieval_enhanced_vlm,Han_CVPR24_rematch_crossmodal_retrieval}, text-to-image generation~\cite{ramesh_2022_dalle2,Rombach_CVPR22_StableDiffusion}, and visual question answering~\cite{Liu_NeurIPS23_LLaVa,Liu_CVPR24_LLaVA2}.

While CLIP achieves remarkable performance in broad domains and tasks, its image and text alignment is still not perfect.
For example, CLIP models tend to encode images and texts into different clusters for each modality, and thus, there is a \textit{modality gap} between images and text features even after sufficiently training with the massive datasets~\cite{Liang_NeurIPS22_clip_modality_gap,Zhang_ICLR23_diagnosing_modality_gap,Qian_NeurIPS24_intra_modal_proxy_clip}.
This modality gap suggests that CLIP has difficulty in precisely mapping images and text.
In fact, \citet{Liang_NeurIPS22_clip_modality_gap} have shown that the modality gap largely affects downstream task performance, especially in fine-grained classification tasks, and \citet{Ray_NeurIPS23_cola_vlm_benchmark} have demonstrated that CLIP models often fail to retrieve localized objects and attributes in images from corresponding captions.\looseness-1

\begin{figure}[t]
    \centering
    \includegraphics[width=1.0\linewidth]{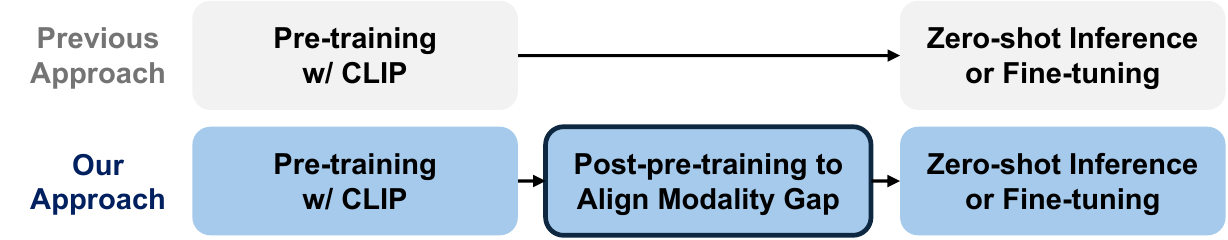}
    \caption{
    Scope of our work: post-pre-training to align the modality gap in pre-trained CLIP models.
    We aim to address the modality gap and enhance the generalization performance of pre-trained CLIP models through lightweight training.
    }
    \label{fig:top}
    \vspace{-5mm}
\end{figure}

To address the modality gap in CLIP, existing literature mainly focuses on refining the CLIP feature spaces in the pre-training or fine-tuning phases.
For pre-training, one approach is multi-task learning of the contrastive objective and auxiliary losses such as self-supervised loss with data augmentations~\cite{Li_ICLR22_declip,Lee_NeurIPS22_uniclip}. 
Another direction for pre-training is to reduce the modality gap by modifying the encoder architectures to explicitly share the weights or feature maps between the image and text encoders~\cite{You_ECCV22_modality_shared_encoder,Chen_CVPR23_FDT}.
For fine-tuning, previous works have demonstrated that the target downstream task performance can be improved significantly by prompt tuning that optimizes trainable visual/textual tokens on target tasks to match image and text in the feature space (i.e., reducing the modality gap)~\cite{Zhou_2022_CoOp,Lu_CVPR22_promptda, Khattak_CVPR23_maple}.
More recently, \citet{Yang_CVPR24_mma} have introduced adapter parameters to share the information across modalities. This cross-modal shared adapter is fine-tuned to solve the target tasks.

\begin{figure*}[t]
    \centering
    \includegraphics[width=0.9\linewidth]{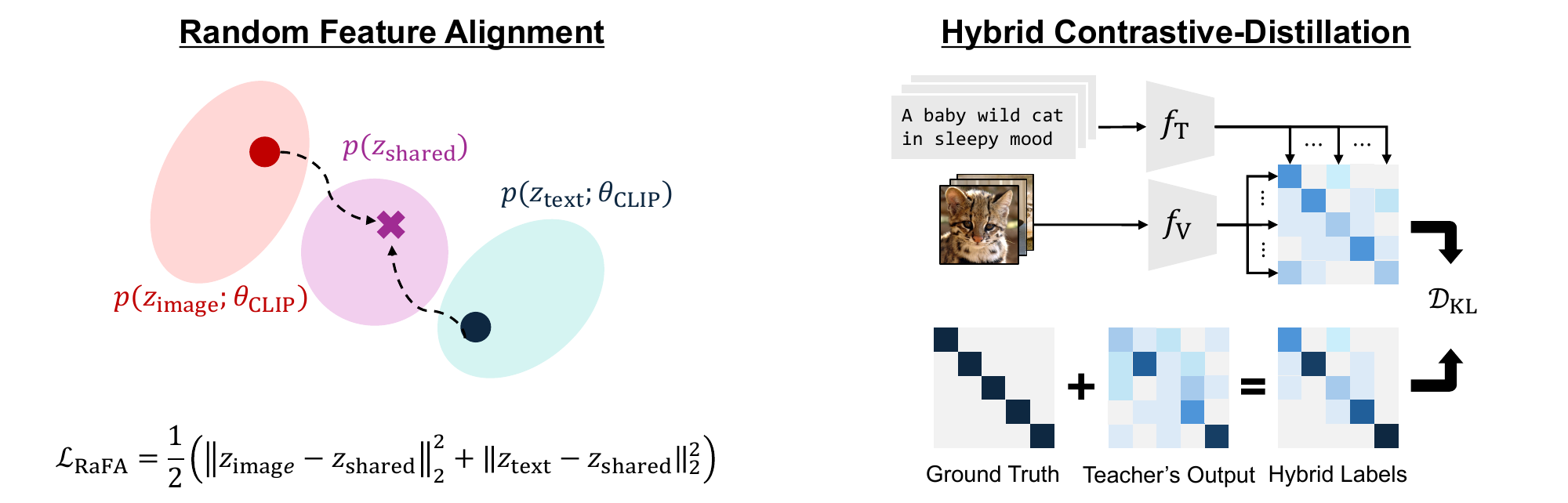}
    \caption{
    Intuition of CLIP-Refine (proposed method). CLIP-Refine modifies a pre-trained CLIP model by \uline{Random Feature Alignment} (RaFA, left) and \uline{Hybrid Contrastive-Distillation} (HyCD, right). RaFA penalizes image and text features to follow a shared prior distribution by minimizing the gap between the features and a random reference vector sampled from the prior. HyCD trains the model with the KL-divergence-based distillation loss using hybrid soft labels composed by alpha blending the ground truth label of image-text pairs with the output of the pre-trained model (teacher).
    }
    \label{fig:method}
    \vspace{-5mm}
\end{figure*}

Although these pre-training/fine-tuning methods succeed in improving the cross-modal alignment capability, they face difficulties in terms of computational cost and zero-shot transfer performance.
On the one hand, pre-training methods require training from scratch with million-scale image-text datasets like LAION-400M~\cite{schuhmann_2021_laion400m}.
Since the previous works often do not publish pre-trained models using such large datasets, na\"ive CLIP models published by OpenAI~\cite{Radford_ICML21_CLIP} or OpenCLIP~\cite{Cherti_CVPR23_openclip} are still a practical choice under the limited budget.
On the other hand, fine-tuning methods do not require huge computation costs because their datasets are much smaller than pre-training ones. 
Nevertheless, fine-tuning pre-trained CLIP models degrades the general zero-shot transfer performance due to focusing on target tasks~\cite{Lai_ICCV23_padclip}.
This is inevitable as the primary goal of fine-tuning is to improve the target task performance.
Therefore, we seek a new approach that is lighter than pre-training and avoids degrading zero-shot transfer performance in fine-tuning.

In this paper, we tackle aligning the image and text features of CLIP models in a training phase called \textit{post-pre-training} between pre-training and fine-tuning (Fig.~\ref{fig:top}).
The goal of post-pre-training is to mitigate the modality gap and enhance zero-shot transfer performance of off-the-shelf pre-trained CLIP models by only using reasonable computation resources and datasets (e.g., a single GPU and COCO Captions~\cite{Lin_2014_COCO}).
This is challenging because na\"ive alignment methods, such as directly minimizing the gap between image and text features, corrupt the feature spaces in terms of the uniformity on the hypersphere~\cite{Wang_ICML20_align_and_uniformity_contrastive}; the uniformity is an important property of contrastive learning representation that indicates the amount of input data information.
Furthermore, post-pre-training with contrastive loss causes catastrophic forgetting of the general knowledge in pre-trained CLIP models due to the overfitting by the restricted mini-batch sizes.\looseness-1

To this end, we introduce a method for post-pre-training called \textit{CLIP-Refine} (Figure~\ref{fig:method}), which is composed of \textit{random feature alignment} (RaFA) and \textit{hybrid contrastive-distillation} (HyCD).
Instead of directly minimizing the gap between features, our idea is to refine the image and text feature distributions to follow a shared prior (e.g., standard Gaussian).
To achieve this, RaFA minimizes the gap between the image/text feature vectors and the reference vectors, which are randomly generated from the prior.
In RaFA, the reference vectors are shared for the pairs of input image and text.
Thus, RaFA penalizes image and text features to explicitly follow the same distribution.
By matching the image and text feature distributions, the features are expected to avoid excessive concentration and preserve uniformity among samples, resulting in a good balance of multi-modal alignment and uniformity in the feature space.
Conversely, HyCD is designed to avoid catastrophic forgetting by modified self-distillation loss with the supervision of image-text pairs.
Specifically, we apply the knowledge distillation where the teacher is the pre-trained model, i.e., minimizing KL-divergence between the outputs of the teacher and (student) post-pre-training model.
Furthermore, we encourage learning new knowledge by blending the teacher's similarity matrix and the identity matrix, which represents the supervision of matched image-text pairs.
Combining RaFA and HyCD, the CLIP models can reduce the modality gap while maintaining the uniformity of features without forgetting past knowledge.

Our extensive experiments on zero-shot classification (12 datasets) and cross-modal retrieval (2 datasets) demonstrate that our method significantly improves zero-shot performance over post-pre-training baselines with contrastive loss.
Through the quantitative and qualitative analysis of feature spaces, we found that CLIP-Refine not only reduces the modality gap but also improves the uniformity of the hypersphere.
This suggests that post-pre-training facilitates effective cross-modal alignment and well-clustered features.

\section{Related Work}\label{sec:relatedwork}
\noindent\textbf{CLIP and Modality Gap.}
CLIP~\cite{Radford_ICML21_CLIP,Jia_ICML21_ALIGN_contrastive} is a multi-modal representation learning technique that trains a model to embed inputs from image and text modalities into shared feature space. This is done by making the correct pairs of image and text closer in the feature space while the other pairs in mini-batch repelled via the InfoNCE-based contrastive loss~\cite{Oord_2018_InfoNCE}.
While this simple multi-modal pre-training has remarkably advanced multi-modal research in various fields~\cite{Xu_EMNLP21_Videoclip,Wang_EMNLP22_medclip,Wang_CVPR22_CLIP_NeRF,Chen_CVPR23_CLIP2Scene,Cheng_CVPR24_Denoising_CLIP,Chen_CVPR24_CLIP_3D_Scene,Wang_CVPR24_CLIP_unsupervised_seg}, \citet{Liang_NeurIPS22_clip_modality_gap} have revealed that pre-trained CLIP models encode image and text into different clusters for each modality, i.e., modality gap, and the small temperature parameter in the CLIP's contrastive loss can cause the modality gap empirically.
The successor work by~\citet{Qian_NeurIPS24_intra_modal_proxy_clip} has theoretically shown that contrastive loss can not reduce the modality gap perfectly.

\noindent\textbf{Pre-training Modifications.}
To mitigate the modality gap and improve the cross-modal alignment, several works have modified the contrastive loss of CLIP by adding auxiliary losses, including geometric cyclic consistency between image and text features~\cite{Goel_NeurIPS22_CyCLIP}, the fine-grained similarity between output tokens of both modalities~\cite{Yao_ICLR22_FILIP}, self-supervised loss with data augmentation~\cite{Li_ICLR22_declip,Lee_NeurIPS22_uniclip}, supervised contrastive learning~\cite{Yang_CVPR22_UniCL}, textual augmented contrastive loss using LLM-generated positive/negative texts~\cite{Doveh_CVPR23_CLIP_text_augmentation}.
Another direction is to explicitly share the information of image and text modalities by introducing shared encoder weights~\cite{You_ECCV22_modality_shared_encoder} or shared feature spaces of finite discrete tokens~\cite{Chen_CVPR23_FDT} to the pre-training.
These methods, however, require large-scale image-text pair datasets (e.g., CC12M~\cite{Changpinyo_CVPR21_CC12M} and LAION-400M~\cite{schuhmann_2021_laion400m}) to achieve practical performance, incurring high computational costs on multiple GPUs.
In contrast, our post-pre-training method improves the generalization performance of off-the-shelf pre-trained models by only using small datasets like Flickr8/30K~\cite{Rashtchian_NAACL10_Flickr} and COCO Caption~\cite{Lin_2014_COCO} with a single GPU.
Besides, our method is cooperative with these pre-training methods because it can be used for any pre-trained model.

\noindent\textbf{Fine-tuning Methods\footnote{We refer to training models on specific target tasks as fine-tuning.}.}
The original paper of CLIP~\cite{Radford_ICML21_CLIP} has reported lower zero-shot performance in several fine-grained classification tasks, such as Aircraft~\cite{maji_13_aircraft}, demonstrating the imperfectness of the cross-modal alignment of CLIP.
Motivated by this, Zhou et al.~\cite{Zhou_2022_CoOp,Zhou_CVPR22_CoCoOp} have proposed to optimize text prompts to mitigate the modality gap by updating additional trainable vectors while the pre-trained models are fixed.
This simple approach helps match image and text in the feature space (i.e., reducing the modality gap) and significantly improves target performance. 
Similarly, The method of \citet{Jia_ECCV22_visual_prompt} learns the additional trainable parameters for image input (visual prompt), and the successor works~\cite{Khattak_CVPR23_maple,Shen_WACV24_multitask_prompt} have unified the visual and textual prompt tuning by the multi-task learning objectives.
Another approach is to directly share the low-rank adaptation parameters across the modalities and update them in fine-tuning~\cite{Yang_CVPR24_mma}.
In the perspective of the refinement of CLIP feature spaces, \citet{Oh_NeurIPS23_m2mix} have theoretically and empirically shown that fine-tuned CLIP models retain the suboptimal feature space in terms of alignment and uniformity, which are important properties to assess the feature quality in contrastive learning~\cite{Wang_ICML20_align_and_uniformity_contrastive}.
To refine the feature space, they have provided a mixup-based fine-tuning method, which generates hard negative samples on the hypersphere in a cross-modal manner ($m^2$-mix).
These fine-tuning techniques succeed in reducing the gap in target datasets, but they do not intend to improve the zero-shot transfer performance.
Our post-pre-training method differs from the fine-tuning methods in the objective: our method aims to improve the zero-shot transfer performance by reducing the modality gap.
Obviously, we can leverage any fine-tuning techniques after post-pre-training by our method, and it helps to boost the baseline performance.\looseness-1

\begin{algorithm}[t]
    \caption{CLIP-Refine for Post-pre-training}\label{alg:clip-refine}
    \begin{algorithmic}[1]
    {\small
        \REQUIRE{Training dataset \(\mathcal{D}\), vision encoder \(f_\mathrm{V}\), text encoder \(f_\mathrm{T}\), pre-trained parameters \(\{\theta_\mathrm{V}^\mathrm{p}, \theta_\mathrm{T}^\mathrm{p}\}\), training batchsize \(B\), step size \(\eta\), temperature parameter \(\tau\), hyper-parameters \(\alpha\)
        \ENSURE{Post-pre-trained parameters \(\{\theta_\mathrm{V}, \theta_\mathrm{T}\}\)}}
        \STATE{\text{\color{gray}\# Initialize parameters}}
        \STATE{\(\theta_\mathrm{V}\leftarrow\theta_\mathrm{V}^\mathrm{p}\)}
        \STATE{\(\theta_\mathrm{T}\leftarrow\theta_\mathrm{T}^\mathrm{p}\)}
        \WHILE{not converged}
        \STATE{\(\{(x^i,t^i)\}^B_{i=1}\sim \mathcal{D}\)}
        \STATE{\(\{z_\mathrm{img}^i, z_\mathrm{txt}^i\}^B_{i=1} \leftarrow \{f_\mathrm{V}(x^i;\theta_\mathrm{V}), f_\mathrm{T}(t^i;\theta_\mathrm{T})\}^B_{i=1}\)}
        \STATE{\(\{z_\mathrm{img}^{i,\mathrm{p}}, z_\mathrm{txt}^{i,\mathrm{p}}\}^B_{i=1} \leftarrow \{f_\mathrm{V}(x^i;\theta_\mathrm{V}^\mathrm{p}), f_\mathrm{T}(t^i;\theta_\mathrm{T}^\mathrm{p})\}^B_{i=1}\)}
        \STATE{\text{\color{gray}\# Compute random feature alignment loss}}
        \STATE{\(\{z_\mathrm{ref}^i | z_\mathrm{ref}^i \sim p(z)\}^B_{i=1}\)}
        \STATE{\(\mathcal{L}_\mathrm{RaFA}\leftarrow \frac{1}{2B}\sum^B_{i=1}\|z^i_\mathrm{img}-z^i_\mathrm{ref}\|^2_2 + \|z^i_\mathrm{txt}-z^i_\mathrm{ref}\|^2_2 \)}
        \STATE{\text{\color{gray}\# Compute hybrid contrastive-distillation loss}}
        \STATE{Compute student outputs~\(p^{\mathrm{I\to T}}_{i,j}\)~and~\(p^{\mathrm{T\to I}}_{i,j}\) for all sample combinations in a batch by Eq.~(\ref{eq:output_i_to_t}) with \(\tau\).}
        \STATE{Compute teacher outputs~\(\hat{q}^{\mathrm{I\to T}}_{i,j}\)~and~\(\hat{q}^{\mathrm{T\to I}}_{i,j}\) for all sample combinations in a batch by Eq.~(\ref{eq:output_hycd_i_to_t}) with \(\tau\) and \(\alpha\).}
        \STATE{\(\mathcal{L}^{\mathrm{I}\to\mathrm{T}}_\mathrm{HyCD} \leftarrow \frac{1}{B}\sum^B_{i=1}\sum^B_{j=1} \hat{q}^{\mathrm{I\to T}}_{i,j}\log\frac{\hat{q}^{\mathrm{I\to T}}_{i,j}}{p^{\mathrm{I\to T}}_{i,j}}\)}
        \STATE{\(\mathcal{L}^{\mathrm{T}\to\mathrm{I}}_\mathrm{HyCD} \leftarrow \frac{1}{B}\sum^B_{i=1}\sum^B_{j=1} \hat{q}^{\mathrm{T\to I}}_{i,j}\log\frac{\hat{q}^{\mathrm{T\to I}}_{i,j}}{p^{\mathrm{T\to I}}_{i,j}}\)}
        \STATE{\(\mathcal{L}_\mathrm{HyCD} \leftarrow \frac{1}{2} (\mathcal{L}^{\mathrm{I}\to\mathrm{T}}_\mathrm{HyCD} + \mathcal{L}^{\mathrm{T}\to\mathrm{I}}_\mathrm{HyCD}) \)}
        \STATE{\(\theta_\mathrm{V} \leftarrow \theta_\mathrm{V} - \frac{\eta}{2}\nabla_{\theta_\mathrm{V}}(\mathcal{L}_\mathrm{RaFA}+\mathcal{L}_\mathrm{HyCD})\)}
        \STATE{\(\theta_\mathrm{T} \leftarrow \theta_\mathrm{T} - \frac{\eta}{2}\nabla_{\theta_\mathrm{T}}(\mathcal{L}_\mathrm{RaFA}+\mathcal{L}_\mathrm{HyCD})\)}
        \ENDWHILE
        }
    \end{algorithmic}
\end{algorithm}

\section{Method}\label{sec:method}
We propose post-pre-training of CLIP in between pre-training and fine-tuning to mitigate the modality gap (Figure~\ref{fig:top}).
To this end, we present CLIP-Refine (Figure~\ref{fig:method}), which is composed of random feature alignment (RaFA) and hybrid contrastive-distillation (HyCD).
RaFA penalizes the model to induce both modal features into a single shared distribution by minimizing the gap between the paired features and a random reference vector.
To reconcile learning new knowledge with retaining past knowledge, HyCD updates the models with knowledge distillation loss using fixed pre-trained models, where the teacher outputs are mixed with ground-truth labels of image-text pairs.
Algorithm~\ref{alg:clip-refine} shows the overall procedure of CLIP-Refine.

\subsection{Problem Setting: Post-pre-training}
We consider a problem setting called post-pre-training, where we aim to improve the cross-modal alignment and generalization performance of pre-trained vision-language models.
This setting allows to access a vision encoder \(f_\mathrm{V}:\mathcal{X}\to\mathbb{R}^{d}\) and a text encoder \(f_\mathrm{T}:\mathcal{T}\to\mathbb{R}^{d}\) with the parameter \(\theta^\mathrm{CLIP}=\{\theta_\mathrm{V}^\mathrm{CLIP},\theta_\mathrm{T}^\mathrm{CLIP}\}\) pre-trained by CLIP, where \(\mathcal{X}\) and \(\mathcal{T}\) are the image and text space.
We optimize the parameters \(\theta_\mathrm{V}\) of \(f_\mathrm{V}\) and \(\theta_\mathrm{T}\) of \(f_\mathrm{T}\) on a post-pre-training image-text pair dataset \(\mathcal{D}=\{(x_i,t_i)\}^{N}_{i=1}\); we basically assume that \(\mathcal{D}\) is much smaller than the pre-training dataset and contains images and text captions in general domain.

\subsection{Objective Function}
In CLIP-Refine, we optimize \(\theta_\mathrm{V}\) and \(\theta_\mathrm{T}\) by the following objective function:
\begin{eqnarray}
     &\min\limits_{\theta_\mathrm{V},\theta_\mathrm{T}}   \mathcal{L}_\mathrm{RaFA}(\theta_\mathrm{V},\theta_\mathrm{T})+ \mathcal{L}_\mathrm{HyCD}(\theta_\mathrm{V},\theta_\mathrm{T}),\label{eq:obj}
\end{eqnarray}
where \(\mathcal{L}_\mathrm{RaFA}\) is the random feature alignment loss and \(\mathcal{L}_\mathrm{HyCD}\) is the hybrid contrastive-distillation loss.
One can consider balancing \(\mathcal{L}_\mathrm{RaFA}\) and \(\mathcal{L}_\mathrm{HyCD}\) by hyperparameters, but we found that their equal contribution achieves the best performance (see Appendix).
We describe the details of \(\mathcal{L}_\mathrm{RaFA}\) and \(\mathcal{L}_\mathrm{HyCD}\) in the following sections.

\subsection{Random Feature Alignment}
The primary goal in post-pre-training is to minimize the modality gap and encourage cross-modal alignment.
To this end, the most straightforward approach is just minimizing the feature gap between the modalities.
Let \(z^i_\mathrm{img}=f_\mathrm{V}(x^i;\theta_\mathrm{V})\) and \(z^i_\mathrm{txt}=f_\mathrm{T}(t^i;\theta_\mathrm{T})\) are the normalized feature vectors of an image-text pair \((x^i, t^i)\).
A na\"ive loss for minimizing the modality gap can be defined by an \(L_2\) distance form as
\begin{equation}\label{eq:naive_gap_loss}
    \mathcal{L}_\mathrm{align} = \mathbb{E}_{(x^i,t^i)\in \mathcal{D}} \|z^i_\mathrm{img} - z^i_\mathrm{txt}\|^2_2.
\end{equation}
However, minimizing this loss function degrades the generalization performance.
This can be explained by the balance of alignment and uniformity ~\cite{Wang_ICML20_align_and_uniformity_contrastive}, which are desirable properties of transferability in contrastive representation learning. 
Alignment is defined by the gap between positive (i.e., image-text) pairs in the feature space, and uniformity is defined by equal separation among all features on the hypersphere.
That is, minimizing Eq.~(\ref{eq:naive_gap_loss}) enhances the alignment of the positive pair but destroys the uniformity on the hypersphere by forcing changes to the feature distribution.
In fact, \citet{Wang_ICML20_align_and_uniformity_contrastive} have shown that using only the alignment loss in post-pre-training degrades the validation performance due to corrupt uniformity.

To address this challenge, we introduce the idea of matching feature distributions instead of directly matching the paired features.
In other words, we refine the multi-modal features of pre-trained CLIP models to follow a prior distribution \(p(z)\) (e.g., standard Gaussian \(\mathcal{N}(0,I)\)) shared across the image and text modalities.
Inspired by random feature regularization in single modal fine-tuning~\cite{Zhong_CVPR20_RandReg,Yamaguchi_CVPR24_AdaRand}, we formulate this idea as minimization between image/text feature vectors \((z^i_\mathrm{img},z^i_\mathrm{txt}\)) and a random reference vector \(z^i_\mathrm{ref}\) sampled from the shared prior \(p(z)\):
\begin{equation}\label{eq:loss_rafa}
    \mathcal{L}_\mathrm{RaFA} = \mathbb{E}_{(x^i,t^i)\in \mathcal{D}} \frac{1}{2} (\|z^i_\mathrm{img} - z^i_\mathrm{ref}\|^2_2 + \|z^i_\mathrm{txt} - z^i_\mathrm{ref}\|^2_2).
\end{equation}
Through this loss function, we can expect that (i) the modality gap is indirectly minimized via \(z^i_\mathrm{ref}\) shared with each image-text pair, (ii) the feature vectors follow the modality shared prior \(p(z)\) gradually~\cite{Yamaguchi_CVPR24_AdaRand}, (iii) the randomness of \(z^i_\mathrm{ref}\) helps to avoid the models from overfitting to \(\mathcal{D}\)~\cite{Zhong_CVPR20_RandReg} and reform the models' capacity~\cite{Shen_CVRP24_stable_rank_tuning}.
Furthermore, we found that \(\mathcal{L}_\mathrm{RaFA}\) can improve the uniformity from baselines (Table~\ref{tab:quantitative_feature_eval}), meaning that the learned representations transfer well to downstream tasks.
Throughout this paper, we use standard Gaussian \(\mathcal{N}(0,I)\) as \(p(z)\) by default; we discuss the effect of prior choice in Sec.~\ref{sec:analysis_prior}.

\setlength{\tabcolsep}{6pt}
\begin{table*}[t]
  \centering
  \caption{
    Zero-shot classification performance on 12 datasets with CLIP ViT-B/32.
   }
  \label{tab:zeroshot-classification}
  \vspace{-3mm}
  \resizebox{1.0\textwidth}{!}{
    \begin{tabular}{lccccccccccccc}
    \toprule
    \multicolumn{1}{l}{\textbf{Method}} & \textbf{Air} & \textbf{Bird} & \textbf{Cal} & \textbf{Car} & \textbf{DTD} & \textbf{Euro} & \textbf{Flo} & \textbf{Food} & \textbf{IN} & \textbf{Pet} & \textbf{SUN} & \textbf{UCF} & \textbf{Avg.} \\
    \midrule
    \multicolumn{1}{l}{Pre-trained} & 18.81 & 49.37 & 23.98 & 60.07 & 40.90 & 38.20 & 59.75 & 80.21 & 59.04 & 81.79 & 60.85 & 59.90 & 52.74 \\\midrule
    \multicolumn{1}{l}{Contrastive} & 13.98 & 40.07 & 23.77 & 47.63 & 36.22 & 34.68 & 46.28 & 70.31 & 52.96 & 70.13 & 54.78 & 58.26 & 45.75 \\
    \multicolumn{1}{l}{$m^2$-mix~\cite{Oh_NeurIPS23_m2mix}} & 14.64 & 40.76 & 23.49 & 49.86 & 36.49 & 34.65 & 47.42 & 71.27 & 53.58 & 70.32 & 56.51 & 58.76 & 46.48 \\
    \multicolumn{1}{l}{Self-KD} & 18.96 & 51.52 & 23.94 & 60.04 & 41.17 & 38.36 & 59.55 & 80.02 & 59.06 & 81.85 & 60.68 & 60.16 & 52.94 \\
    \multicolumn{1}{l}{HyCD} & 18.72 & 50.50 & 23.98 & 59.83 & 40.48 & 40.90 & 58.06 & 80.19 & 59.71 & 81.45 & 62.04 & 61.70 & 53.13 \\
    \multicolumn{1}{l}{HyCD + $\mathcal{L}_\mathrm{align}$} & 13.29 & 34.20 & 23.73 & 43.70 & 35.00 & 39.27 & 46.27 & 73.31 & 52.57 & 72.42 & 55.01 & 58.55 & 45.61 \\
    \rowcolor{blue!20}  
    \multicolumn{1}{l}{CLIP-Refine (Ours)} & \textbf{20.77} & \textbf{52.72} & \textbf{24.79}& \textbf{60.63} & \textbf{41.49} & \textbf{41.44} & \textbf{62.64} & \textbf{81.87} & \textbf{60.93} & \textbf{83.57} & \textbf{63.63} & \textbf{61.85} & \textbf{54.69} \\
    \bottomrule
    \end{tabular}}
  \vspace{-3mm}
\end{table*}%

\subsection{Hybrid Contrastive-Distillation}
Another important goal of post-pre-training is to retain the past knowledge in pre-trained CLIP models.
To this end, our basic strategy is to penalize the post-pre-training models with knowledge distillation loss~\cite{hinton_2015_distilling} where the frozen pre-trained CLIP model performs as the teacher.
We define the mini-batch-wise knowledge distillation loss with KL-divergence according to the formulation of CLIP-KD~\cite{Yang_CVPR24_CLIP-KD} as 
\begin{eqnarray}
    &\mathcal{L}_\mathrm{KD} = \frac{1}{2}(\mathcal{L}^\mathrm{I\to T}_\mathrm{KD} + \mathcal{L}^\mathrm{T\to I}_\mathrm{KD})\label{eq:naive_kd},\\
    &\mathcal{L}^\mathrm{I\to T}_\mathrm{KD} = \frac{1}{B}\sum^B_{i=1}\sum^B_{j=1} q^{\mathrm{I\to T}}_{i,j}\log\frac{q^{\mathrm{I\to T}}_{i,j}}{p^{\mathrm{I\to T}}_{i,j}}\label{eq:loss_kd_i_to_t},\\
    &p^{\mathrm{I\to T}}_{i,j} = \frac{\exp(z^i_\mathrm{img}\cdot z^j_\mathrm{txt}/\tau)}{\sum^B_{k=1}\exp(z^i_\mathrm{img}\cdot z^k_\mathrm{txt}/\tau)}\label{eq:output_i_to_t},\\
    &q^{\mathrm{I\to T}}_{i,j} = \frac{\exp(z^{i,\mathrm{p}}_\mathrm{img}\cdot z^{j,\mathrm{p}}_\mathrm{txt}/\tau)}{\sum^B_{k=1}\exp(z^{i,\mathrm{p}}_\mathrm{img}\cdot z^{k,\mathrm{p}}_\mathrm{txt}/\tau)}\label{eq:output_pre_i_to_t},
\end{eqnarray}
where \(\tau\) is the temperature parameter, \((z^{i,\mathrm{p}}_\mathrm{img}, z^{i,\mathrm{p}}_\mathrm{txt}) = (f_\mathrm{V}(x^i;\theta^\mathrm{CLIP}_\mathrm{V}),f_\mathrm{T}(t^i;\theta^\mathrm{CLIP}_\mathrm{T}))\) are the feature vectors generated by the pre-trained image/text encoders, and \(\mathcal{L}^\mathrm{T\to I}_\mathrm{KD}\) is defined by swapping the order of image features and text features in Eq.~(\ref{eq:output_i_to_t})~and~(\ref{eq:output_pre_i_to_t}).
Although minimizing Eq.~(\ref{eq:naive_kd}) retains the past knowledge, we found that it interferes with RaFA because it strongly enchains the models' parameters to the pre-trained ones.

To reconcile RaFA and the retention of past knowledge, we present a technique that blends the ground-truth label of an image-text pair with the teacher output.
Concretely, we modify Eq.~(\ref{eq:output_pre_i_to_t}) by alpha-blending of an indicator function \(\mathbb{I}_{i=j}\), which returns one if \(i=j\) otherwise zero, and the teacher output signal:
\begin{equation}\label{eq:output_hycd_i_to_t}
    \hat{q}^{\mathrm{I\to T}}_{i,j} = \alpha \mathbb{I}_{i=j} + (1-\alpha) q^{\mathrm{I\to T}}_{i,j},
\end{equation}
where \(\alpha\) is a hyperparameter to balance acquiring new knowledge and retaining past knowledge; we use \(\alpha=0.5\) by default and discuss the effect of \(\alpha\) in the Appendix.
With this modification, the post-pre-training models can further enhance the cross-modal alignment by learning the correct image-text pair (i.e., \(i=j\)) with reference to the relative knowledge contained in the \(i\neq j\) coordinates of the teacher's outputs.
By using \(\hat{q}^{\mathrm{I\to T}}_{i,j}\) and \(\hat{q}^{\mathrm{T\to I}}_{i,j}\), we compute the HyCD loss function as follows:
\begin{eqnarray}
    &\mathcal{L}_\mathrm{HyCD} = \frac{1}{2}(\mathcal{L}^\mathrm{I\to T}_\mathrm{HyCD} + \mathcal{L}^\mathrm{T\to I}_\mathrm{HyCD})\label{eq:loss_hycd},\\
    &\mathcal{L}^\mathrm{I\to T}_\mathrm{HyCD} = \frac{1}{B}\sum^B_{i=1}\sum^B_{j=1} \hat{q}^{\mathrm{I\to T}}_{i,j}\log\frac{\hat{q}^{\mathrm{I\to T}}_{i,j}}{p^{\mathrm{I\to T}}_{i,j}}\label{eq:loss_hycd_i_to_t}.
\end{eqnarray}
While \(\mathcal{L}_\mathrm{HyCD}\) solely improves the generalization performance of the pre-trained CLIP models, combining it with \(\mathcal{L}_\mathrm{RaFA}\) achieves larger improvements by minimizing the modality gaps without losing the past knowledge.

\section{Experiments}\label{sec:experiments}
We evaluate CLIP-Refine on 12 classification datasets and 2 cross-modal retrieval datasets by using multiple pre-trained models.
We also conduct qualitative and quantitative analyses to assess the feature space through modality gap, alignment, uniformity metrics, and PCA visualization.

\subsection{Settings}\label{sec:ex_setting}
\noindent\textbf{Baselines.}
We compare our CLIP-Refine with the following baselines of post-pre-training.
\uline{\textbf{Pre-trained}}: predicting with pre-trained weights without additional training.
\uline{\textbf{Contrastive}}: post-pre-training with the contrastive loss~\cite{Radford_ICML21_CLIP,Jia_ICML21_ALIGN_contrastive}.
\uline{\textbf{\(m^2\)-mix}}~\cite{Oh_NeurIPS23_m2mix}: post-pre-training by the modified contrastive loss with multi-modal mixup; we use this fine-tuning method as a baseline because it can be used for refining the feature spaces as suggested in the original paper~\cite{Oh_NeurIPS23_m2mix}.
\uline{\textbf{Self-KD}}: using Eq.~(\ref{eq:naive_kd}) for post-pre-training.
\uline{\textbf{HyCD}}: using only Eq.~(\ref{eq:loss_hycd}) for training.
\uline{\textbf{HyCD+\(\mathcal{L}_\mathrm{align}\)}}: combining Eq.~(\ref{eq:loss_hycd}) and Eq.~(\ref{eq:naive_gap_loss}).

\noindent\textbf{Post-pre-training Datasets.}
We used COCO Caption~\cite{Lin_2014_COCO} as the default post-pre-training dataset.
We also used Flickr8K/30K~\cite{Rashtchian_NAACL10_Flickr}, CC3M~\cite{Sharma_ACL18_CC3M}, and CC12M~\cite{Changpinyo_CVPR21_CC12M} to confirm the dataset size effect in post-pre-training.

\noindent\textbf{Test Datasets.}
We used 12 image classification datasets containing various image domains: Aircraft (Air)~\cite{maji_13_aircraft}, Bird~\cite{Welinder_10_cub2002011}, Caltech-101 (Cal)~\cite{FeiFei_caltech101} Car~\cite{krause_3DRR2013_stanford_cars}, DTD~\cite{cimpoi_CVPR14_DTD}, EuroSAT (Euro)~\citep{Helber_IEEE_eurosat}, Flower (Flo)~\cite{Nilsback_08_flowers}, Food~\cite{bossard14_Food101}, ImageNet (IN)~\cite{russakovsky_imagenet}, Pet~\cite{parkhi_CVPR12_oxford_pets}, SUN397~\cite{Xiao_CVPR10_sun397}, and UCF-101~\cite{Soomro_arXiv12ucf101}.
We also evaluated our method on COCO2017-Val and Flickr8K/30K for the image and text retrieval tasks; we denote text-to-image retrieval as $\mathrm{T\to I}$ and image-to-text retrieval as $\mathrm{I\to T}$.
We adopt these datasets since they are often used to evaluate the zero-shot transfer performance of CLIP models~\citep{Zhou_2022_CoOp}.
We evaluated models on the test sets except for ImageNet, and the official validation set for ImageNet.

\noindent\textbf{Pre-trained Models.}
By default, we used CLIP-ViT-B/32~\cite{Radford_ICML21_CLIP} with the pre-trained weights downloaded from the OpenAI's official repositories.
We also tested other pre-trained models, including OpenCLIP-ViT-H/14, OpenCLIP-ViT-bigG/14, SigLIP~\cite{Zhai_ICCV23_SigLIP} and DFN~\cite{Fang_ICLR24_DFN} with the pre-trained weights on the OpenCLIP~\cite{Cherti_CVPR23_openclip} repository.

\noindent\textbf{Post-pre-training.}
In all settings, we trained models by the AdamW~\cite{Loshchilov_ICLR19_AdamW} optimizer with the learning rate of 1.0\(\times\)10\(^{-6}\) for one epoch; we determined the learning rate by the zero-shot classification validation using the subset of the ImageNet training set constructed by uniform sampling.
We used mini-batch sizes of 512 as the default.
For CLIP-Refine, we used \(\alpha=0.5\).
The input samples were preprocessed using the default image transformations provided by each pre-trained model.
We implemented the training and evaluation with PyTorch-1.13.
We ran the experiments on a 24-core Intel Xeon CPU with a single NVIDIA A100 GPU with 80GB VRAM.\looseness-1

\setlength{\tabcolsep}{6pt}
\begin{table*}[t]
  \centering
  \caption{
    Zero-shot retrieval performance on COCO2017-Val and Flickr8K/30K with CLIP ViT-B/32.
   }
  \label{tab:zeroshot-retrieval}
  \vspace{-3mm}
  \resizebox{1.0\textwidth}{!}{
    \begin{tabular}{l ccc>{\columncolor{gray!10}}c>{\columncolor{gray!10}}c>{\columncolor{gray!10}}c ccc>{\columncolor{gray!10}}c>{\columncolor{gray!10}}c>{\columncolor{gray!10}}c ccc>{\columncolor{gray!10}}c>{\columncolor{gray!10}}c>{\columncolor{gray!10}}c}
    \toprule
    & \multicolumn{6}{c}{\uline{\textbf{COCO2017-Val}}} & \multicolumn{6}{c}{\uline{\textbf{Flickr8K}}} & \multicolumn{6}{c}{\uline{\textbf{Flickr30K}}}\\
    & \multicolumn{3}{c}{\textbf{T$\to$I}} & \multicolumn{3}{c}{\cellcolor{gray!10}\textbf{I$\to$T}} & \multicolumn{3}{c}{\textbf{T$\to$I}} & \multicolumn{3}{c}{\cellcolor{gray!10}\textbf{I$\to$T}} & \multicolumn{3}{c}{\textbf{T$\to$I}} & \multicolumn{3}{c}{\cellcolor{gray!10}\textbf{I$\to$T}} \\
    \multicolumn{1}{l}{\textbf{Method}} & \textbf{R@1} & \textbf{R@5} & \textbf{R@10} &\textbf{R@1} & \textbf{R@5} & \textbf{R@10} & \textbf{R@1} & \textbf{R@5} & \textbf{R@10} &\textbf{R@1} & \textbf{R@5} & \textbf{R@10} & \textbf{R@1} & \textbf{R@5} & \textbf{R@10} & \textbf{R@1} & \textbf{R@5} & \textbf{R@10} \\
    \midrule
    \multicolumn{1}{l}{Pre-trained} & 30.56 & 54.92 & 65.26 & 33.26 & 59.10 & 68.78 & 30.06 & 53.83 & 63.29 & 32.42 & 55.57 & 65.26 & 21.06 & 40.96 & 50.53 & 24.52 & 44.57 & 54.14 \\\midrule
    \multicolumn{1}{l}{Contrastive} &  34.88 & 61.50 & 72.10 & 31.86 & 56.80 & 67.48 &  34.94 & 59.41 & 69.36 & 28.83 & 51.55 & 61.07 & 25.69 & 47.32 & 56.93 & 18.42 & 36.29 & 45.37 \\
    \multicolumn{1}{l}{$m^2$-mix~\cite{Oh_NeurIPS23_m2mix}} & 36.28 & 62.18 & 73.08 & 32.92 & 57.96 & 67.56 & 35.04 & 59.66 & 69.66 & 28.64 & 50.65 & 60.77 & 25.99 & 47.18 & 57.19 & 19.88 & 37.63 & 46.51 \\
    \multicolumn{1}{l}{Self-KD} & 31.04 & 55.58 & 65.90 & 35.36 & 59.00 & 69.72 & 29.98 & 53.83 & 63.55 & 32.47 & 55.11 & 65.10 & 21.92 & 41.51 & 50.75 & 24.94 & 44.82 & 54.27 \\
    \multicolumn{1}{l}{HyCD} & 36.04 & 62.28 & 73.14 & 37.88 & 62.54 & 72.84 & 35.72 & 60.34 & 70.14 & 34.69 & 57.19 & 67.22 & 27.11 & 49.12 & 58.78 & 27.17 & 47.66 & 56.94 \\
    \multicolumn{1}{l}{HyCD + $\mathcal{L}_\mathrm{align}$} & 33.92 & 61.18 & 72.06 & 35.14 & 61.74 & 72.06 & 31.73 & 55.63 & 65.90 & 32.58 & 56.94 & 66.99 & 23.78 & 44.73 & 54.47 & 23.58 & 45.07 & 55.23 \\
    \rowcolor{blue!20}  
    \multicolumn{1}{l}{CLIP-Refine (Ours)} & \textbf{37.64} & \textbf{63.54} & \textbf{74.42} & \textbf{38.78} & \textbf{65.04} & \textbf{75.12} & \textbf{36.11} & \textbf{61.29} & \textbf{70.51} & \textbf{38.28} & \textbf{61.55} & \textbf{70.94} & \textbf{26.65} & \textbf{49.70} & \textbf{59.03} & \textbf{30.99} & \textbf{51.54} & \textbf{61.31} \\
    \bottomrule
    \end{tabular}}
  \vspace{-3mm}
\end{table*}%

\paragraph{Evaluation Metrics.}
For zero-shot transfer performance, we report top-1 accuracy in classification tasks and recall@$k$ scores in cross-modal retrieval tasks; recall@$k$ indicates the proportion of correct answers in the retrieved top-$k$ candidates for all test input samples. 
We denote recall@$k$ as R@$k$.
To quantitatively evaluate the post-pre-trained features, we measure modality gap, alignment, and uniformity scores.
Modality gap~\cite{Liang_NeurIPS22_clip_modality_gap} assesses a cluster-wise gap between image and text features:
\begin{align}
  \text{Modality~Gap} := \|\bar{f}_\mathrm{V}-\bar{f}_\mathrm{T}\|^2_2,
\end{align}
where \(\bar{f}_\cdot\) is a mean feature vector for all test samples of each modality; lower is better.
Alignment~\cite{Wang_ICML20_align_and_uniformity_contrastive} is defined by 
\begin{align}
  \text{Alignment} := \frac{1}{N_\mathrm{test}}\sum^{N_\mathrm{test}}_{i=1} \|f_\mathrm{V}(x^i) - f_\mathrm{T}(t^i)\|^2_2,
\end{align}
representing how well the positive pair is aligned in the feature space; lower is better.
Uniformity~\cite{Wang_ICML20_align_and_uniformity_contrastive} evaluates how uniformly distributed the feature vectors are on the hypersphere, which is defined by radial basis function (RBF) kernel as 
\begin{align}
  \text{Uniformity} := \frac{1}{2N_\mathrm{test}}\sum_{f_1,f_2\in F}\exp({-2\|f_1 - f_2\|^2_2}),
\end{align}
where \(F = \{f_\mathrm{V}(x^i)\}^{N_\mathrm{test}}_{i=1} \cup\{f_\mathrm{T}(t^i)\}^{N_\mathrm{test}}_{i=1}\); lower is better.
We report the average score for each metric by running experiments three times.

\subsection{Zero-shot Transfer Evaluation}\label{sec:exp_zero-shot_transfer}
We first demonstrate the zero-shot performance of our CLIP-Refine.
Tables~\ref{tab:zeroshot-classification}~and~\ref{tab:zeroshot-retrieval} list the zero-shot test performance on classification and cross-modal retrieval tasks after post-pre-training.
For zero-shot classification tasks, post-pre-training by the contrastive loss (i.e., Contrastive and $m^2$-mix) largely degraded the top-1 accuracy performance for all datasets.
This can be caused by catastrophic forgetting since Self-KD did not show performance degradation.
One cause of catastrophic forgetting is a smaller batchsize than at pre-training; the OpenAI's CLIP uses a batchsize of 32,768 with hundreds of GPUs~\cite{Radford_ICML21_CLIP}, whereas our setting uses ones of 512 with a single GPU.
Since a smaller batch contains a limited number of negative samples, contrastive learning tends to overfit and catastrophically forget past knowledge.
We discuss more details about the effect of batchsize in Sec.~\ref{sec:analsys_batchsize}.
In contrast, our CLIP-Refine impressively improved the zero-shot performance for all datasets.
The ablation studies of HyCD show that the performance gain comes from the combination of HyCD with RaFA.
Notably, HyCD+\(\mathcal{L}_\mathrm{align}\) resulted in the worst performance among the baselines, suggesting that na\"ively minimizing the gap between image and text features leads to the degradations of generalization performance. 

For zero-shot retrieval, our CLIP-Refine achieved the best performance with a larger margin than that in classification.
Interestingly, in contrast to classification tasks, we found that Contrastive and $m^2$-mix improved the pre-trained models in the $\mathrm{T\to I}$ retrieval tasks while the $\mathrm{I\to T}$ performance was degraded.
This indicates that the image encoders are more prone to overfitting than the text encoders.
This can be explained by the difference between image and text in the data spaces; images are represented by continuous-valued pixels and thus have a high degree of freedom, while texts are represented by discrete tokens in vocabulary and thus have a limited degree of freedom.
Therefore, the image encoder tends to overfit to a limited number of negative samples in small batches, and the text encoder can learn new knowledge from them while keeping the past knowledge.
Since the zero-shot classification is a $\mathrm{I\to T}$ retrieval task, this can also be a reason for the performance degradation in classification.
In this perspective, our CLIP-Refine successfully prevents the overfitting in $\mathrm{I\to T}$ and greatly improves the performance in both $\mathrm{I\to T}$ and $\mathrm{T\to I}$.\looseness-1

\begin{figure*}[t]
    \centering
        \begin{tabular}{cccc}
          \begin{minipage}[t]{0.24\textwidth}
                \centering
                \includegraphics[align=c,width=\linewidth]{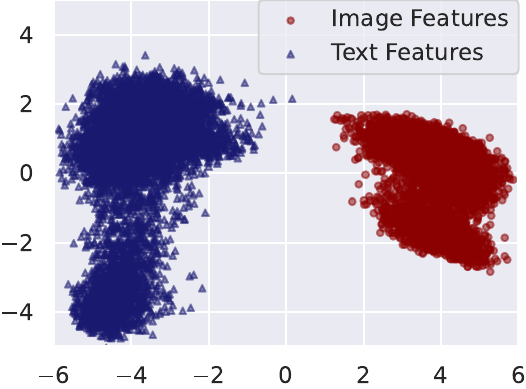}\subcaption{Pre-trained} 
          \end{minipage}&
          \begin{minipage}[t]{0.24\textwidth}
                \includegraphics[align=c,width=\linewidth]{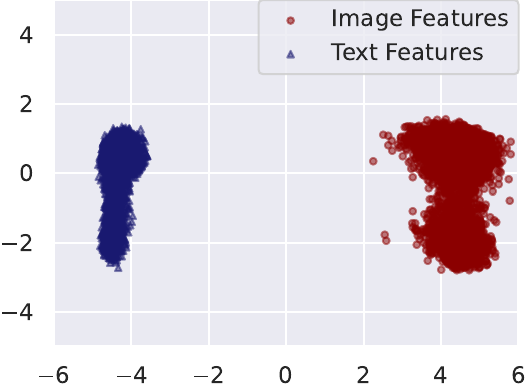}\subcaption{Contrastive}
          \end{minipage}&
          \begin{minipage}[t]{0.24\textwidth}
                \includegraphics[align=c,width=\linewidth]{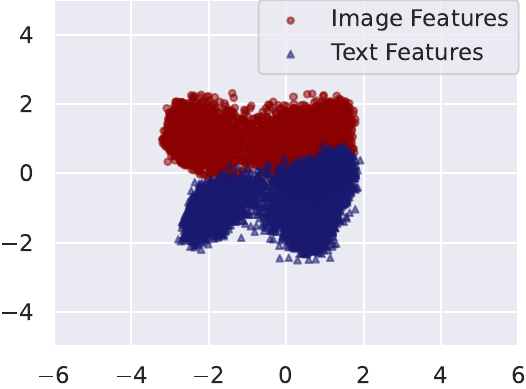}\subcaption{HyCD+\(\mathcal{L}_\mathrm{align}\)}
          \end{minipage}&
          \begin{minipage}[t]{0.24\textwidth}
                \includegraphics[align=c,width=\linewidth]{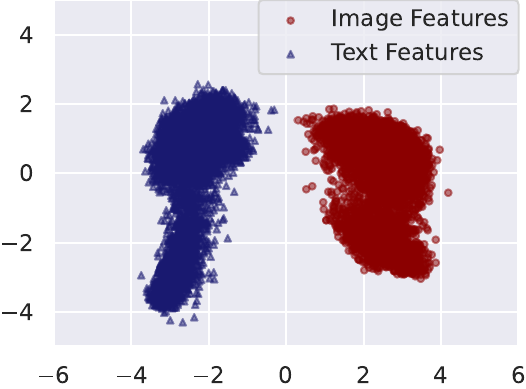}\subcaption{CLIP-Refine (Ours)}
          \end{minipage}\\
        \end{tabular}
        \caption{
            PCA visualization of multi-modal feature spaces of trained models (Flickr8K, ViT-B/32).
            CLIP-Refine reduces the modality gap while keeping the form of feature distributions for each modality.
            }
        \label{fig:feature_viz}
        \vspace{-3mm}
 \end{figure*}
 
\subsection{Analysis of Feature Space}
In this section, we aim to answer our fundamental research question: how much can we alleviate the modality gap by post-pre-training with CLIP-Refine?
For quantitative evaluations, we summarize the modality gap, alignment, and uniformity scores in Table~\ref{tab:quantitative_feature_eval}.
HyCD+\(\mathcal{L}_\mathrm{align}\) achieved the best modality gap and alignment scores. 
However, it significantly degraded the uniformity score, indicating that directly minimizing \(\mathcal{L}_\mathrm{align}\) can destroy feature space well constructed by the CLIP training.
In contrast, our CLIP-Refine consistently improved all of the scores.
This suggests that it is important to minimize not only the gap between image and text features but also the uniformity for improving the generalization performance of pre-trained CLIP models, which is consistent with the findings in~\cite{Wang_ICML20_align_and_uniformity_contrastive}.
In such a sense, since CLIP-Refine penalizes the image and text features to follow the shared reference distribution \(p(z)=\mathcal{N}(0,I)\), where the samples tend to be uniformly distributed on the hypersphere~\cite{PRML_Bishop}, it can naturally achieve a good balance of alignment and uniformity, and reasonably helps to reduce the modality gap.

\begin{table}[t]
    \centering
    \caption{Quantitative evaluation of features (Flickr8K, ViT-B/32).}
    \label{tab:quantitative_feature_eval}
    \resizebox{1.0\columnwidth}{!}{
    \begin{tabular}{lccc}
        \toprule
        Method & Modality Gap ${\times 10^{-3}}$ ($\downarrow$) & Alignment ($\downarrow$) & Uniformity ($\downarrow$) \\
        \midrule
        Pre-trained & 1.3345 & 1.3724 & 0.0895 \\
        Contrastive & 2.1714 & 1.4300 & 0.2136 \\
        $m^2$-mix~\cite{Oh_NeurIPS23_m2mix} & 2.0911 & 1.4130 & 0.2019 \\
        Self-KD & 1.2754 & 1.3774 & 0.0901 \\
        HyCD & 1.3823 & 1.3578 & 0.0971 \\
        HyCD+$\mathcal{L}_\mathrm{align}$ & \textbf{0.1590} & \textbf{0.7413} & 0.1172 \\
        \rowcolor{blue!20}  
        CLIP-Refine (Ours) & 0.7934 & 1.2849 & \textbf{0.0495} \\
        \bottomrule
    \end{tabular}
    }
    \vspace{-4mm}
\end{table}

As a qualitative evaluation, Figure~\ref{fig:feature_viz} illustrates the PCA visualization of post-pre-trained features for each method.
We can see that the contrastive model makes the modality gap rather large.
HyCD+\(\mathcal{L}_\mathrm{align}\) succeeds in gathering the image and text features into the same cluster but distorts the feature distribution for each modality.
Meanwhile, CLIP-Refine certainly reduces the modality gap while keeping the shape of the pre-trained features.
Although the modality gap is not completely zero, it is natural because there remains a constant value in the lower bound of the modality gap~\cite{Qian_NeurIPS24_intra_modal_proxy_clip}.

\subsection{Detailed Analysis}
\subsubsection{Prior Distributions}\label{sec:analysis_prior}

Here, we analyze CLIP-Refine by varying the prior distribution \(p(z)\).
According to the prior works~\cite{Zhong_CVPR20_RandReg,Yamaguchi_CVPR24_AdaRand}, we tried standard Gaussian \(\mathcal{N}(0,I)\), uniform distribution \(U(0,1)\), Gaussian distributions using pre-trained text feature statistics \(\mathcal{N}(\mu^\mathrm{p}_\mathrm{txt},(\sigma^\mathrm{p}_\mathrm{txt})^2)\), image feature statistics \(\mathcal{N}(\mu^\mathrm{p}_\mathrm{img},(\sigma^\mathrm{p}_\mathrm{img})^2)\), statistics of all image and text features \(\mathcal{N}(\mu^\mathrm{p}_\mathrm{all},(\sigma^\mathrm{p}_\mathrm{all})^2)\), where \(\mu^\mathrm{p}_\cdot, \sigma^\mathrm{p}_\cdot\) is mean and variance of image or/and text feature vectors computed on a pre-trained CLIP model.
We also tested variations of \(\mathcal{N}(0,\beta I)\) by varying $\beta$ in \(\{0, 0.01, 0.1, 10, 100\}\); \(\beta=0\) means RaFA without randomness, i.e.,  \(\frac{1}{2}(\|z^i_\mathrm{img}\|^2_2+\|z^i_\mathrm{txt}\|^2_2)\).

\begin{table}[t]
    \centering
    \caption{Evaluation of CLIP-Refine varying prior distributions in \(\mathcal{L}_\mathrm{RaFA}\). ZS Cls. is the zero-shot classification accuracy averaged on 12 datasets.}
    \label{tab:prior_choice}
    \resizebox{0.9\columnwidth}{!}{
    \begin{tabular}{lccc}
        \toprule
        Prior & ZS Cls. & Alignment & Uniformity \\\midrule
        Pretrained & 52.74 & 1.3724 & 0.0895 \\
        \midrule
        \(U(0,1)\) & 52.91 & 1.3081 & 0.0661 \\
        \(\mathcal{N}(\mu^\mathrm{p}_\mathrm{img},(\sigma^\mathrm{p}_\mathrm{img})^2)\) & 52.93 & 1.3142 & 0.0916 \\
        \(\mathcal{N}(\mu^\mathrm{p}_\mathrm{txt},(\sigma^\mathrm{p}_\mathrm{txt})^2)\) & 53.40 & 1.3304 & 0.0873 \\
        \(\mathcal{N}(\mu^\mathrm{p}_\mathrm{all},(\sigma^\mathrm{p}_\mathrm{all})^2)\) & 53.29 & 1.3361 & 0.0888 \\
        \(\mathcal{N}(0,0I)\) & 53.79 & 1.3052  & 0.0554 \\
        \(\mathcal{N}(0,0.01I)\) & 54.19 & 1.3079 & 0.0544 \\
        \(\mathcal{N}(0,0.1I)\) & 54.32 & 1.3021 & 0.0515 \\
        \(\mathcal{N}(0,I)\) & \textbf{54.69} & \textbf{1.2849} & \textbf{0.0495} \\
        \(\mathcal{N}(0,10I)\) & 54.23 & 1.3085 & 0.0537 \\
        \(\mathcal{N}(0,100I)\) & 53.59 & 1.3441 & 0.0663 \\
        \bottomrule
    \end{tabular}
    }
    \vspace{-4mm}
\end{table}

Table~\ref{tab:prior_choice} shows the zero-shot classification performance, alignment, and uniformity scores when varying \(p(z)\).
Using the uniform distribution \(U(0,1)\) did not show large improvements on the zero-shot performance and uniformity score.
This may be because random reference vectors from \(U(0,1)\) do not necessarily concentrate on the hypersphere in contrast to \(\mathcal{N}(0,I)\), possibly causing the collapse of uniformity by \(\mathcal{L}_\mathrm{RaFA}\).
In this perspective, we recommend using \(\mathcal{N}(0,I)\) or families of Gaussian distributions.
Interestingly, among the Gaussian distributions with the pre-trained statistics, the prior with the text feature, i.e., \(\mathcal{N}(\mu^\mathrm{p}_\mathrm{txt} (\sigma^\mathrm{p}_\mathrm{txt})^2)\) largely improved the zero-shot performance.
This indicates that aligning image features with text features is effective in zero-shot classification tasks and also evidences the performance improvements in the \(\mathrm{I}\to \mathrm{T}\) tasks observed in Sec.~\ref{sec:exp_zero-shot_transfer}.
More importantly, the ablation studies of the variance magnitude \(\beta\) in the zero-mean Gaussian distributions emphasize the importance of randomness in \(\mathcal{L}_\mathrm{RaFA}\).

\subsubsection{Post-pre-training Datasets}\label{sec:analsys_post-pre-training_datasets}
We evaluate the effects of the dataset choice in post-pre-training.
Table~\ref{tab:multiple_datasets} shows the zero-shot classification/retrieval performance of CLIP-Refine with Flickr8K/30K, COCO Caption, CC3M, and CC12M as post-pre-training datasets.
We confirm that most of the CLIP-Refine models outperformed the pre-trained baselines.
Notably, larger datasets do not always achieve high performance; CC3M and CC12M are inferior to COCO Caption.
This suggests that the quality of image-text pairs in a dataset is an important factor in post-pre-training.
Since CC3M and CC12M contain mismatched and noisy image-text pairs~\cite{Han_CVPR24_rematch_mismatched_pair}, COCO Caption that has higher-quality text captions can be superior in terms of the modality alignment of pre-trained CLIP models.
In fact, when we filtered out low-quality image-text pairs by CLIP-Score-based filtering of DataComp~\cite{datacomp}, the models achieved significant performance improvements.
This indicates that the caption quality is crucial and our method is scalable if we have high-quality image-text pairs.

\begin{table}[t]
    \centering
    \caption{Zero-shot classification (12 datasets) and retrieval (Flicker30K) performance of CLIP-Refine with various post-pre-training datasets with ViT-B/32.}
    \label{tab:multiple_datasets}
    \resizebox{\columnwidth}{!}{
    \begin{tabular}{lcccc}
        \toprule
        & & \uline{Zero-shot Cls. Acc.} & \multicolumn{2}{c}{\uline{Zero-shot Ret. R@5}}\\
        Post-pre-training Dataset & \# of Samples & Avg. of 12 datasets & T $\rightarrow$ I & I $\rightarrow$ T \\
        \midrule
        None (Pre-trained) & 0 & 52.74 & 40.96 & 44.57 \\\midrule
        Flickr8K & 8K & 53.04 & 45.57 & 46.80 \\
        Flickr30K & 30K & 53.74 & 48.66 & 49.25 \\
        COCO Caption & 118K & 54.69 & 49.70 & 51.54 \\
        CC3M & 3M & 53.80 & 49.56 & 50.24 \\
        CC12M & 12M & 52.46 & 45.35 & 47.79 \\
        \midrule
        Filtered CC3M & 1M & 55.50 & 52.45 & 53.00 \\
        Filtered CC12M & 4M & 60.12 & 54.97 & 57.88 \\
        \bottomrule
    \end{tabular}
    }
\end{table}

\subsubsection{Pre-trained Models}\label{sec:analsys_pretrained_models}
We investigate the generalizability of CLIP-Refine to pre-trained CLIP models.
Table \ref{tab:model_performance_comparison} shows the zero-shot performance with 6 pre-trained models.
We observed that our CLIP-Refine stably improved the pre-trained baseline.
This suggests that the bigger and stronger models still have the modality gap, and modifying the gap by CLIP-Refine helps improve the generalization performance for various pre-trained models.

\begin{table}[t]
    \centering
    \caption{Zero-shot classification (12 datasets) and retrieval (Flickr30K) performance with various pre-trained CLIP models.}
    \label{tab:model_performance_comparison}
    \resizebox{1.0\columnwidth}{!}{
    \begin{tabular}{lcccccc}
        \toprule
              & \multicolumn{3}{c}{\uline{Zero-shot Cls. Acc.}} & \multicolumn{3}{c}{\uline{Zero-shot Ret. R@5 (T$\to$I / I$\to$T)}}\\
        Model & Pre-trained & Contrastive & CLIP-Refine & Pre-trained & Contrastive & CLIP-Refine \\
        \midrule
        ViT-B/32 & 52.74 & 45.76 & \textbf{54.69} & 40.96/44.57 & 47.32/36.29 & \textbf{49.70}/\textbf{51.54}\\
        ViT-L/14 & 62.96 & 61.90 & \textbf{63.88} & 49.13/52.77 & 51.07/51.00 & \textbf{58.38}/\textbf{58.41}  \\
        OpenCLIP ViT-H/14 & 68.52 & 57.61 & \textbf{69.05} & 65.84/66.02 & 57.47/43.43 & \textbf{67.82}/\textbf{68.48} \\
        OpenCLIP ViT-G/14 & 71.28 & 57.44 & \textbf{72.59} & 68.73/68.38 & 66.35/42.36 & \textbf{70.94}/\textbf{69.76} \\
        SigLIP ViT-SO400M/14 & 72.57 & 65.63 & \textbf{73.61} & 72.77/72.83 & 64.63/62.97 & \textbf{76.54}/\textbf{76.49} \\
        DFN ViT-H/14 & 77.30 & 76.39 & \textbf{78.29} & 71.93/70.89 & 72.81/70.09 & \textbf{76.39}/\textbf{75.44} \\
        \bottomrule
    \end{tabular}
    }
    \vspace{-5mm}
\end{table}

\subsubsection{Mini-Batch Sizes}\label{sec:analsys_batchsize}
\begin{figure}[t]
    \centering
    \includegraphics[width=0.7\linewidth]{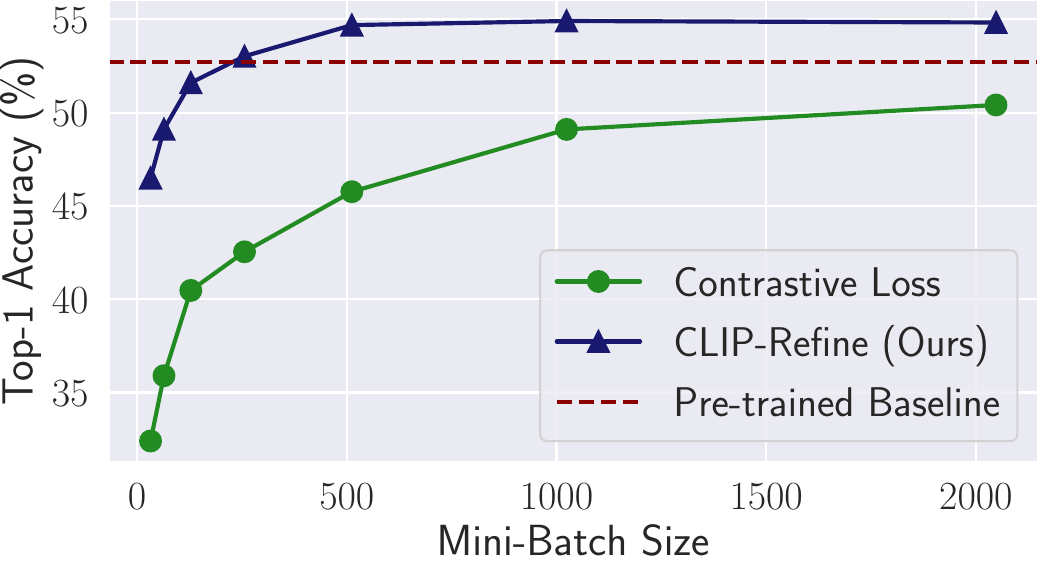}
    \caption{
    Zero-shot classification accuracy averaged on 12 datasets when varying mini-batch sizes in post-pre-training.
    }
    \label{fig:batchsize}
    \vspace{-3mm}
\end{figure}

We provide the analysis of varying mini-batch sizes in post-pre-training.
We tried the mini-batch sizes in \(\{32,64,128,256,512,1024,2048\}\).
We used gradient accumulation by following the OpenCLIP~\cite{Cherti_CVPR23_openclip} implementation for the mini-batch sizes of \(1024\) and \(2048\).
Figure~\ref{fig:batchsize} shows the zero-shot performance versus mini-batch sizes.
The contrastive loss gradually improved the accuracy according to the mini-batch size.
However, the accuracy curve implies that it is difficult to exceed the pre-trained baseline with feasible mini-batch sizes.
This is probably due to the large discrepancy from the mini-batch size in the pre-training (i.e., 32,768), which causes the model to forget knowledge in the predictions for negative pairs in a mini-batch.
In contrast, CLIP-Refine always outperformed the contrastive loss and succeeded in improving the pre-trained baseline with feasible mini-batch sizes.
This suggests that CLIP-Refine's HyCD prevents knowledge forgetting even for small mini-batch sizes with few negative pairs and is able to modify the feature space with new knowledge.

\begin{table}[t]
    \centering
    \caption{Fine-tuning top-1 accuracy on the ImageNet validation set.}
    \label{tab:finetuning}
    \resizebox{0.8\columnwidth}{!}{
    \begin{tabular}{lccc}
        \toprule
        Model & Pre-trained & Contrastive & CLIP-Refine \\
        \midrule
        ViT-B/32 & 73.40 & 69.46 & \textbf{74.14} \\
        \bottomrule
    \end{tabular}
    }
    \vspace{-5mm}
\end{table}

\subsubsection{Effects on Fine-tuning}\label{sec:analysis_finetuning}
We show the transferability of CLIP-Refine to fine-tuning.
We fine-tuned the linear classification head on ImageNet while freezing post-pre-trained models.
Tabel~\ref{tab:finetuning} shows the results.
CLIP-Refine outperformed the pre-trained and contrastive loss baselines.
This emphasizes that our method can generate useful representation even in fine-tuning.

\section{Conclusion}\label{sec:conclusion}
This paper presented CLIP-Refine, a post-pre-training method for pre-trained CLIP models to align the modality gap between the image and text features.
CLIP-Refine addresses the modality gap by penalizing the multi-modal features to follow a shared prior distribution by minimizing the distance to the random reference vectors sampled from the prior.
To maintain the past knowledge in the CLIP models and promote the feature alignment simultaneously, CLIP-Refine also trains the model with knowledge distillation loss using hybrid soft labels composed of ground-truth image-text pair labels and outputs from the pre-trained CLIP model.
Through extensive experiments, we show that CLIP-Refine can improve the zero-shot performance of the pre-trained CLIP by addressing the modality gap and enhancing uniformity.
We believe that our work not only provides a practical method but also opens up a new research field where we refine the pre-trained CLIP models by post-pre-training with much smaller computational costs than that of pre-training.

\clearpage
{\small
\bibliographystyle{ieeenat_fullname}
\bibliography{ref}
}

\onecolumn
\appendix
\renewcommand\thetable{\Roman{table}}
\renewcommand\thefigure{\Roman{figure}}

\begin{figure*}[h]
    \centering
    \begin{minipage}{0.49\linewidth}
    \centering
    \includegraphics[width=0.8\columnwidth]{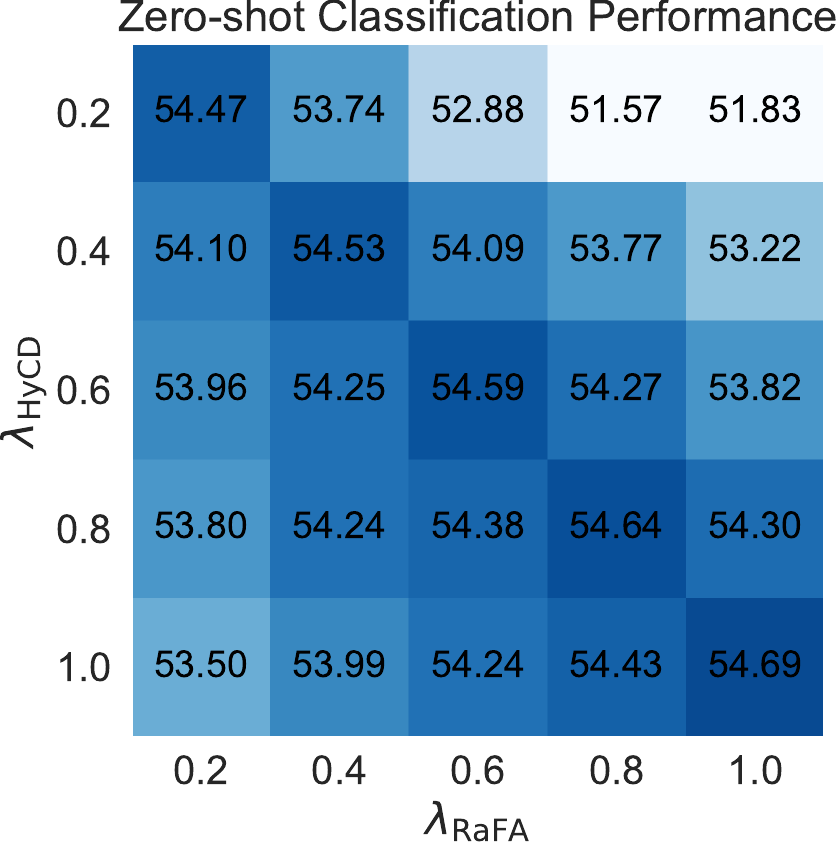}
    \caption{
        Zero-shot classification accuracy averaged on 12 datasets when varying balancing parameters between \(\mathcal{L}_\mathrm{RaFA}\) and \(\mathcal{L}_\mathrm{HyCD}\) (ViT-B/32).
    }
    \label{fig:lambdas}
    \end{minipage}
    \hfill
    \begin{minipage}{0.49\linewidth}
    \centering
    \includegraphics[width=\columnwidth]{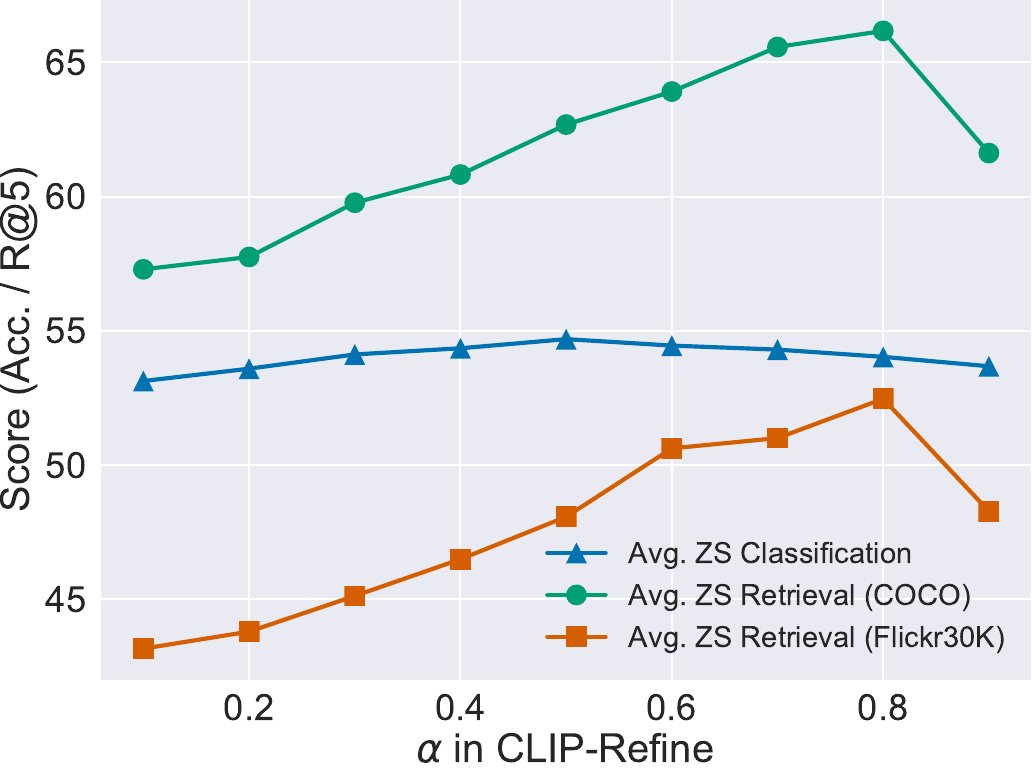}
    \caption{
        Zero-shot performance on 12 classification datasets and retrieval datasets when varying \(\alpha\) in \(\mathcal{L}_\mathrm{HyCD}\) (ViT-B/32).
    }
    \label{fig:alpha}
    \end{minipage}
\end{figure*}

\begin{figure}[t]
    \centering
    \includegraphics[width=0.5\linewidth]{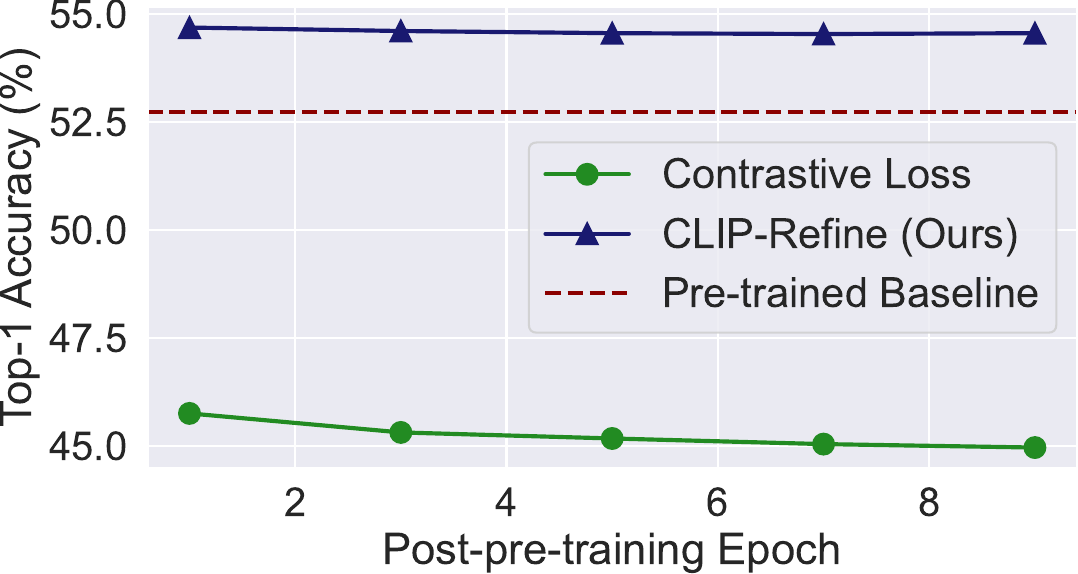}
    \caption{
    Zero-shot classification accuracy averaged on 12 datasets when varying epochs in post-pre-training.
    }
    \label{fig:epoch}
\end{figure}

\section{Effects of Hyperparameters}\label{sec:effect_hyperparam}
In the main paper, we fixed the contributions of \(\mathcal{L}_\mathrm{RaFA}\),  \(\mathcal{L}_\mathrm{HyCD}\) in CLIP-Refine and the hyperparameter of \(\alpha\) in Eq.~(1) for HyCD, and epochs for post-pre-training.
Here, we confirm the effects of varying them on the performance.
\paragraph{Trade-off between \(\mathcal{L}_\mathrm{RaFA}\) and \(\mathcal{L}_\mathrm{HyCD}\)}
We evaluate balancing \(\mathcal{L}_\mathrm{RaFA}\) and \(\mathcal{L}_\mathrm{HyCD}\) in Eq~(1) by introducing hyperparameters \(\lambda_\mathrm{RaFA}\) and \(\lambda_\mathrm{HyCD}\) as follows:
\begin{align}
     \min\limits_{\theta_\mathrm{V},\theta_\mathrm{T}}   \lambda_\mathrm{RaFA}\mathcal{L}_\mathrm{RaFA}(\theta_\mathrm{V},\theta_\mathrm{T})+ \lambda_\mathrm{HyCD}\mathcal{L}_\mathrm{HyCD}(\theta_\mathrm{V},\theta_\mathrm{T}).\nonumber
\end{align}
We varied \(\lambda_\mathrm{RaFA}\) and \(\lambda_\mathrm{HyCD}\) in \(\{0.2,0.4,0.6,0.8,1.0\}\) and post-pre-trained CLIP ViT-B/32 on COCO Caption.
Figure~\ref{fig:lambdas} illustrates the heatmap where each cell represents the zero-shot classification accuracy averaged on 12 datasets.
We can see that the diagonal elements of the heatmap achieve higher performance, indicating that keeping the equal contribution of \(\lambda_\mathrm{RaFA}\) and \(\lambda_\mathrm{HyCD}\) is important for better zero-shot performance.

\paragraph{Trade-off parameter \(\alpha\) in \(\mathcal{L}_\mathrm{HyCD}\)}
We evaluate the trade-off parameter \(\alpha\) in Eq.(8) for balancing learning of the new knowledge from post-pre-training and retaining of the past knowledge in the pre-trained CLIP models.
We varied \(\alpha\) in \(\{0.1, 0.2, 0.3, 0.4, 0.5, 0.6, 0.7, 0.8, 0.9, 1.0\}\).
Figure~\ref{fig:alpha} shows the trend of the averaged zero-shot classification and retrieval accuracy.
We see that the trends in classification and retrieval are different; the classification performance is less sensitive than the retrieval performance, and an overly high value of \(\alpha\) degrades both performances. 
This suggests that prioritizing new knowledge is important but balancing the new and past knowledge is crucial to achieve the best performance.

\paragraph{Post-pre-training Epochs}
We show the effect of increasing post-pre-training epochs from one, which is used in the main paper.
Figure~\ref{fig:epoch} shows the averaged zero-shot classification accuracy when varying the post-pre-training epoch in \(\{1,3,5,7,9\}\).
CLIP-Refine stably kept performance even when increasing epochs, while the contrastive loss slightly degraded the performance according to the epochs.
This implies that our CLIP-Refine can provide stable performance improvements by avoiding catastrophic forgetting even in longer epochs.
This also means that our CLIP-Refine has the practical advantage of not having to search for the appropriate epoch length in each case.

\section{Additional Experiments}
\begin{table}[t]
    \begin{minipage}{0.48\linewidth}
    \caption{\footnotesize Robustness Evaluation on Zero-shot Classification.}
    \label{tb:robustness}
    \vspace{-2mm}
        \resizebox{1.0\columnwidth}{!}{
        \begin{tabular}{lc|cccc}\toprule
            Method & IN1K & V2 & A & R & Sketch\\
          \midrule
            Pre-trained & 59.04 & 51.80 & 28.84 & 64.81 & 38.38   \\
            Contrastive & 37.04 & 45.52 & 22.92 & 62.80 & 35.57   \\
            $m^2$-mix & 59.06 & 46.32 & 22.51 & 63.42 & 35.59   \\
            Self-KD & 51.88 & 52.01 & 28.65 & 65.08 & 38.52   \\
            HyCD+$\mathcal{L}_\mathrm{Align}$ & 57.06 & 45.41 & 21.45 & 62.00 & 34.73   \\
            CLIP-Refine (Ours) & \textbf{60.92} & \textbf{53.51} & \textbf{30.68} & \textbf{67.05} & \textbf{41.46}   \\
            \bottomrule
        \end{tabular}
        }
        \end{minipage}
        \centering
    \begin{minipage}{0.48\linewidth}
    \centering
        \includegraphics[width=\columnwidth]{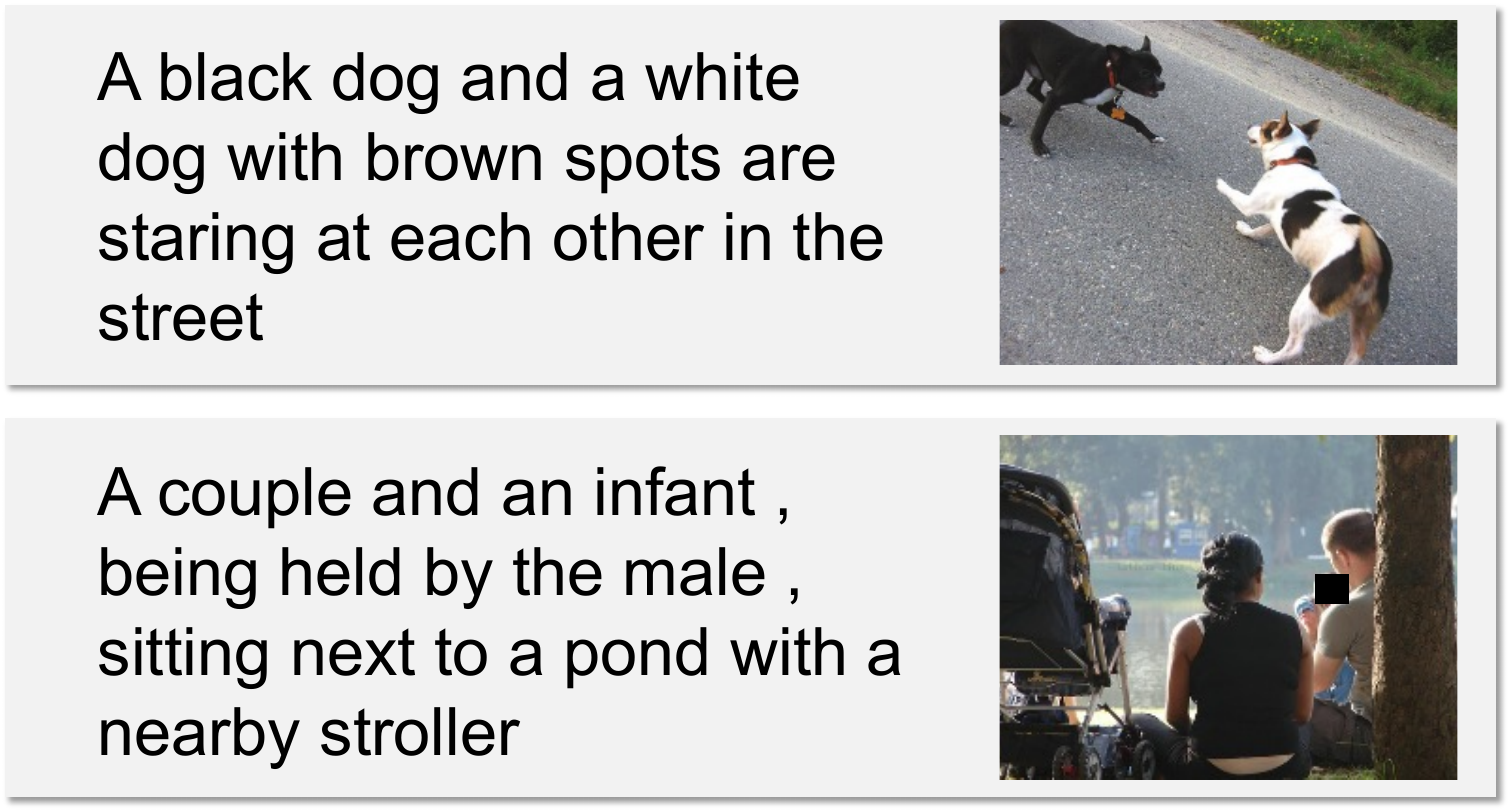}
        \vspace{-7mm}
    \captionof{figure}{Correctly retrieved samples}\label{fig:viz}
    \end{minipage}
\end{table}

\subsection{Robustness Evaluation}
Here, we evaluate the robustness of our method through the evaluation on ImageNet variants including ImageNet-V2~\cite{Recht_ICML19_imagenetv2}, ImageNet-A~\cite{Hendrycks_CVPR21_imageneta}, ImageNet-R~\cite{Hendrycks_ICCV21_imagenetr}, and ImageNet-Sketch~\cite{Wang_NeurIPS19_imagenetsketch}.
Table~\ref{tb:robustness} that our method robustly performs on these variants, supporting the general performance improvements of our method.

\subsection{Visualization Study}
We randomly selected samples of Flickr30K from which CLIP failed, but CLIP-Refine succeeded (Fig.~\ref{fig:viz}).
We see that CLIP-Refine can match complex text and image pairs with multiple attributes and object combinations. 
This highlights that the multi-modal alignment is enhanced by reducing the modality gap.


\end{document}

%% file: preamble.tex
\usepackage[dvipsnames]{xcolor}
\usepackage[utf8]{inputenc} 
\usepackage[T1]{fontenc}    
\usepackage{url}            
\usepackage{booktabs}       
\usepackage{amsfonts}       
\usepackage{amsmath}
\usepackage{nicefrac}       
\usepackage{microtype}      
\usepackage{graphicx}
\usepackage{graphbox}
\usepackage{subcaption}
\usepackage{algorithm}
\usepackage{algorithmic}
\usepackage{bm}
\usepackage[accsupp]{axessibility} 
\usepackage{ulem}
\usepackage{colortbl,array}
\usepackage{multirow}

\newcommand{\TODO}[1]{}
\renewcommand{\TODO}[1]{{\color{cyan} [TODO: {#1}]}}
\newcommand{\WIP}[1]{}
\renewcommand{\WIP}[1]{{\color{magenta} [WIP: {#1}]}}
\definecolor{midblue}{rgb}{0,0.11372549,0.258823529}